# Imagining and building wise machines: The centrality of AI metacognition

**Samuel G. B. Johnson[1*], Amir-Hossein Karimi[2], Yoshua Bengio[3], Nick Chater[4], Tobias Gerstenberg[5], Kate Larson[6], Sydney Levine[7], Melanie Mitchell[8], Iyad Rahwan[9], Bernhard Schölkopf[10], Igor Grossmann[1*]**

[1] University of Waterloo, Department of Psychology
[2] University of Waterloo, Department of Electrical and Computer Engineering
[3] Université de Montréal, Department of Computer Science and Operations Research
[4] Warwick Business School, Behavioural Science Group
[5] Stanford University, Department of Psychology
[6] University of Waterloo, Cheriton School of Computer Science
[7] Google DeepMind
[8] Santa Fe Institute
[9] Max Planck Institute for Human Development
[10] Max Planck Institute for Intelligent Systems

*** Correspondence to**: Sam Johnson (samuel.johnson@uwaterloo.ca) *or*
Igor Grossmann (igrossma@uwaterloo.ca)

## Highlights

- We examine the why and the how of building wise AI.

- Wisdom helps humans to navigate intractable problems through object-level strategies (for managing problems) and metacognitive strategies (for managing object-level strategies).

- Wise metacognition includes strategies such as intellectual humility, perspective-taking, and context-adaptability.

- Wise AI, through such improved metacognitive strategies, would be more robust to new environments, explainable to users, cooperative in pursuing shared goals, and safe in avoiding both prosaic and catastrophic failures.

- We suggest several approaches to benchmarking wisdom, training wise reasoning strategies, and adapting AI architecture for metacognition.

## Abstract

Although AI has become increasingly smart, its wisdom has not kept pace. In this article, we examine what is known about human wisdom and sketch a vision of its AI counterpart. We introduce human wisdom as strategies for solving intractable problems—those outside the scope of analytic techniques—including both 'object-level' strategies like heuristics (for managing problems) and 'metacognitive' strategies like intellectual humility, perspective-taking, or context-adaptability (for managing object-level-task fit). We argue that AI systems particularly struggle with this type of metacognition. Wise metacognition would lead to AI more robust to novel environments, explainable to users, cooperative with others, and safer by risking fewer misaligned goals with human users. We discuss how wise AI might be benchmarked, trained, and implemented.

# Imagining and building wise machines: The centrality of AI metacognition

## Where does AI still struggle?

Despite recent breakthroughs, artificial intelligence systems (AIs) still face critical shortcomings. They struggle in novel, unpredictable environments, lacking **robustness** (see Glossary). Their computations are opaque, creating a problem of **explainability** [1]. Their challenges with communication and credibility create barriers to **cooperation** [2]. These shortcomings limit our ability to harness the benefits of AI while avoiding risks and ensuring **safety** [3]. As AIs increasingly act as agents in the world, these problems will be exacerbated.

But AIs are not the only intelligent agents that must solve these problems—we humans also face analogues of each of them. Might our own solutions yield some clues for how AIs might do so more effectively?

We argue that one core set of capabilities underlies humans' ability to make robust decisions, explain our reasoning, achieve goals cooperatively, and interact safely—**wisdom**. We examine the function and mechanisms of human wisdom, concluding that wisdom serves to solve intractable problems and proceeds via a suite of complementary object-level strategies (which provide possible solutions to problems) and perspectival metacognitive strategies (which are necessary to decide among the solutions). We then consider how humans use these mechanisms to solve our versions of the robustness, explainability, cooperation, and safety problems. By analogy, we suggest that fostering wisdom in AIs—particularly wise metacognition—will help address these problems.

## What is wisdom?

Consider these examples of human wisdom:

- Willa's children are bitterly arguing about money. Willa draws on her life experience to explain why they should instead compromise in the short term and prioritize their sibling relationship in the long term.
- Daphne is a world-class cardiologist. Nonetheless, she consults with a junior colleague when she recognizes that he knows more about a patient's history than she does.
- Ron is a political consultant who formulates possible scenarios to ensure his candidate will win. He not only imagines best case scenarios, but also imagines that his client has lost the election and considers what might have caused the loss.

Why do we intuit some abilities (applying life experience, being intellectually humble, reflective scenario planning) as 'wise,' but not others (solving tricky integrals, cracking clever jokes, composing beautiful sonnets)? Accounts of wisdom highlight a wide array of characteristics [4-10; Box 1]. In our view, differences across theories mask important generalizations about wisdom's function and mechanisms [11].

**Box 1: Wisdom and Metacognition**

Though philosophers have debated wisdom for millennia, empirically-grounded models are recent [4–10]. The Berlin Wisdom Model defines wisdom as expertise in important and difficult life matters, combining knowledge (e.g., about human nature) with certain metacognitive strategies that are sensitive to context, value pluralism, and uncertainty [6]. The MORE Model highlights how wise people build psychological resources—such as managing uncertainty and developing open reflectiveness toward experiences and perspectives—to cope with life's challenges [7]. Balance Theory emphasizes how wise people deploy their knowledge and skills toward the common good by balancing interests (theirs, others', society's) and time horizons (short- and long-term) [10].

Emerging consensus models integrate these perspectives, either conceptually [8] or by surveying wisdom researchers directly [4]. Across approaches, wisdom converges on a cluster of metacognitive skills—context-sensitivity, intellectual humility, interest-balancing, and perspective-integration—which we term **perspectival metacognition**. Rooted in philosophical perspectivism, it shifts the goal of reasoning from finding a single "correct" answer toward achieving a state of maximal situational clarity attained by evaluating and coordinating competing interpretations.

Although individuals vary in these skills, most people show them to some extent, for example when planning ahead or coordinating within social groups [4]. This view challenges the notion that wisdom is reserved for a rare elite; instead, most humans exhibit moments both of wisdom and of folly [9].

Not all metacognition is perspectival. Whereas some metacognitive strategies (e.g., monitoring memory, checking reasoning) optimize performance on well-structured tasks with clear accuracy criteria, perspectival metacognition specifically concerns multiple, often incommensurable perspectives. Recruited for ill-structured, value-laden social problems in which multiple, partly incompatible standpoints must be coordinated rather than simply judged as right or wrong, this subset of metacognition moves beyond egocentric reasoning toward balancing interests, adapting to context, and recognizing epistemic limits when decisions affect others.

Although metacognition is central to wisdom, it does not exhaust it: most wisdom models also treat concern for others and the common good as central components [4,8]. One possibility is that, in many real-world contexts, such as repeated interactions with the same partners [89] or when one's reputation matters to third parties [90], the most effective way to deal with difficult life challenges, even from a self-interested standpoint, *is* to prioritize the common good.

### ***The function of human wisdom: Navigating intractable situations***

If we lived in a textbook, we would not need wisdom. All problems would have correct answers and the world would advertise the information required to find them. Natural selection would have made us nothing more or less than master statisticians, merciless optimizers, lightning calculators. Indeed, in some domains—like low-level visual processing—we approximate this ideal.

Yet, social interaction and decision-making in an unstructured, ever-changing world require further tools [12]. Such problems are often **intractable** in one or more ways. This can happen because of ambiguities in goals—conflicting values that cannot be put on the same scale (*incommensurable* [13]) or a potential outcome changes underlying preferences (*transformative* [14]). It can happen because probabilities cannot be assigned to possible outcomes—the outcomes cannot be enumerated (*radically uncertain* [15]), there is a strong dependency on initial conditions (*chaotic* [16]), the underlying process is changing over time (*non-stationary*), or the situation is far beyond experience (*out-of-distribution*). And it can happen if the optimal outcome is calculable only with infeasible resources (*computationally explosive*).

Our earlier examples of wisdom featured such intractability. Wisdom helped Willa understand how to make an incommensurable trade-off, Daphne to navigate an out-of-distribution situation, and Ron to make useful forecasts despite his ignorance about the radically uncertain future.

***Mechanisms of human wisdom: Metacognitive strategy selection***

We argue that wisdom manages intractable problems by cultivating and deploying two types of strategies [4,11] (Figures 1-2): **Object-level strategies** to manage the problem itself (i.e., the "object" of judgment) and a cluster of **metacognitive strategies** to manage those object-level strategies, particularly when they conflict [17-18].

Object-level strategies yield candidate solutions to intractable problems. Many object-level strategies are **heuristics**—rules of thumb which rely on a small number of inputs and do not attempt a complex analysis [19] but may approximate it [20]. For example, Willa and Ron may have used heuristics like "Prioritize family relationships" and "Avoid the worst-case scenario." Heuristics often work well, despite requiring less computation than optimization, because they focus on just the most relevant information, reducing the chances of overfitting [19]. Much of "folk wisdom" comprises culturally-evolved heuristics, transmitted across generations (e.g., deference to elders).

The trouble with object-level strategies is their multiplicity. Heuristics can conflict ("look before your leap" vs "he who hesitates is lost"), and other classes of strategies, varying in computational complexity, co-exist with heuristics. In **narrative** thinking, a reasoner uses causal knowledge and analogies to construct a mental model that can explain a situation, generate predictions, and evaluate choices [12,21], as when Ron draws on his knowledge and experiences to generate worst-case scenarios. These intuitive strategies also co-exist with **decision technologies**, such as explicit analytic strategies, as when Daphne uses risk-scoring algorithms as one input to clinical decision-making. Wisdom requires us not just to *have* these strategies, but to effectively *manage* them.

Even a well-tailored suite of object-level strategies falls short of wisdom. First, even simple strategies depend on information; an **input-seeking process** is required. (Ron must check if he has the relevant facts for his scenarios and fill any gaps.) Second, strategies often yield conflicting advice; a **conflict resolution process** is required to select the best strategy for each situation [23]. (Should Daphne follow the strategy "trust your judgment" or "trust knowledgeable experts"?) Third, strategies can break under unfavorable conditions, as when the underlying pattern changes unpredictably; an **outcome-monitoring process** is required to safeguard against nonsensical outcomes. (Willa would question her usual advice if one child was taking advantage of the other.)

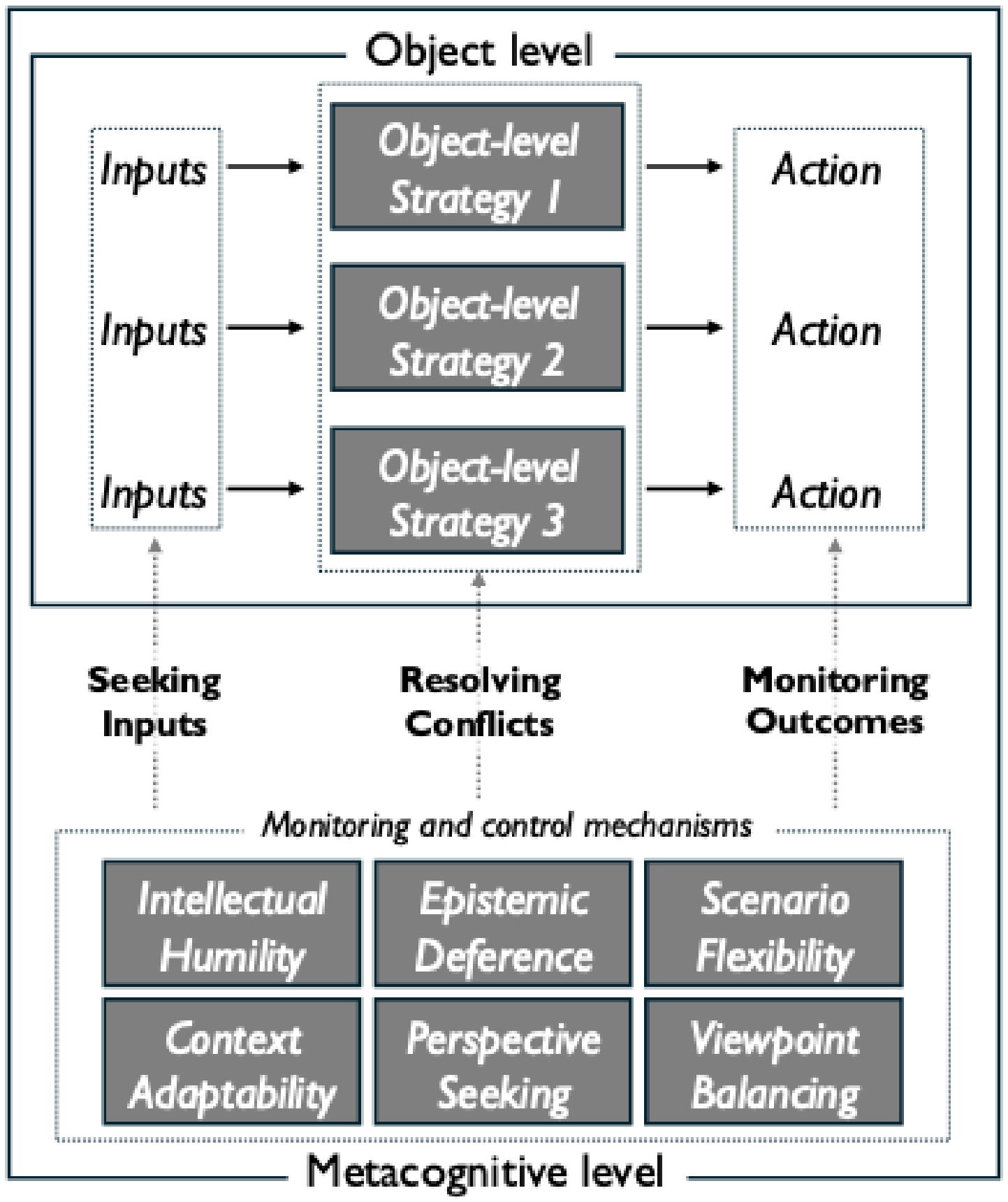

**Figure 1.** The relationship between object-level and metacognitive strategies in wise reasoning. Object-level strategies (e.g., heuristics, narratives, decision technologies) provide candidate actions for a given situation. Metacognitive monitoring and control processes regulate these strategies in three ways: obtaining the appropriate inputs, deciding which strategy to use when they conflict, and monitoring their outcomes to avoid catastrophic actions. (***Key figure.***)

Navigating this complexity requires the ability to monitor and adapt object-level strategies [24-26] using **perspectival metacognition** [4]—strategies for coordinating perspectives, including one's own and others' (Box 1). Some are primarily epistemic: Intellectual humility comprises awareness of what one does (not) know [27]; scenario flexibility involves considering the diverse ways that a scenario could unfold; context adaptability identifies features of a situation that makes it comparable or distinct from other situations [6]. Others have a social dimension: epistemic deference is a willingness to defer to others' expertise [28]; perspective seeking draws on multiple perspectives [6]; viewpoint balancing recognizes and integrates discrepant interests [10].

Perspectival metacognition contributes to the input-seeking, conflict resolution, and outcome-monitoring required to manage object-level strategies (Figure 1). For example, perspective seeking is important for gaining relevant inputs, context adaptability is crucial for resolving conflicts among strategies in a context-sensitive way, and viewpoint balancing is one component of outcome-monitoring. Often, these strategies work together. Daphne exhibits intellectual humility when she recognizes that she does not understand her patient's symptoms (recognizing that her existing object-level strategies are inappropriate); perspective-seeking when she calls upon her colleague's expertise (seeking out new object-level strategies); context adaptability when she considers whether her patient's unique situation limits the relevance of her colleague's expertise (assessing the relevance of new object-level strategies); and ultimately epistemic deference when she adopts her colleague's view (accepting the outcome of the new proposed strategy).

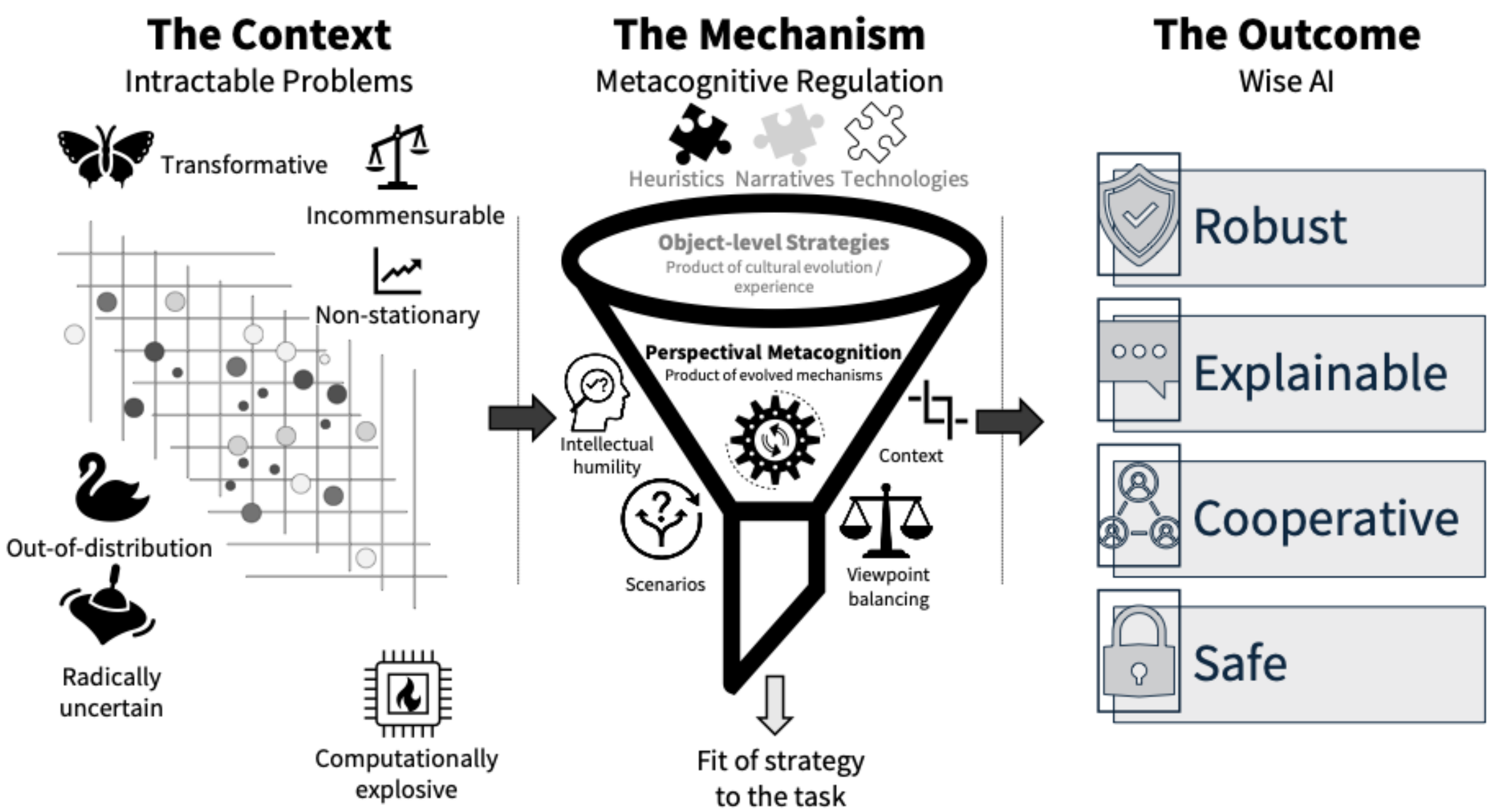

**Figure 2.** From Intractability Problems to Wise AI. ***Left Panel***. Wisdom functions specifically to solve intractable problems—situations that resist analytic optimization due to incommensurable values, radical uncertainty, or non-stationary environments. ***Center Panel***. To address these challenges, agents deploy object-level strategies (e.g., intuitive strategies like heuristics and narratives, as well as technologies like formal procedures) derived from experience or cultural evolution. However, these strategies are often insufficient because they conflict or lack necessary inputs. Perspectival metacognition serves as the regulatory control, applying epistemic strategies (e.g., intellectual humility: awareness of what one does and does not know; acknowledgment of uncertainty and one's fallibility) and social strategies (e.g., viewpoint balancing: recognizing and integrating discrepant interests) to select and adapt the correct object-level approach to the task at hand. ***Right Panel***: Implementing this metacognitive architecture enables AI to move beyond a narrow view of intelligence as optimization toward wisdom, resulting in systems that are more robust, explainable, cooperative, and safe.

### *Toward wise AI: Machine metacognition*

Object-level strategies define the search space, whereas perspectival metacognition provides the tools for its wise navigation (Figure 2). Here, we focus here on the latter because it has been the subject of comparatively less research (but see Box 2 regarding recent advances in AI metacognition broadly). We believe that perspectival metacognition is therefore the current weak link for wise AI. Here, we focus on GenAI systems such as Large Language Models (LLMs), but the arguments extend to other AI paradigms.

GenAI models do have rudimentary forms of metacognition [29]. They can monitor and control some of their neural activations in "neurofeedback" paradigms [30]. They can classify math problems by solving procedure [31] and, on easier problems, can assess whether a step taken led in the correct direction [32]. They perform well on tests of situational competence for relatively unambiguous situations [33], and some models can use an inference-time search to decide when to stop searching. At the same time, they struggle at other metacognitively loaded tasks [29]. They often "hallucinate" an answer rather than admit ignorance [34] and they struggle to understand their goals [35], capabilities [35], and strength of their evidence [36]. This cluster of epistemic failings has been argued to be symptomatic of a broader "metacognitive myopia" [37].

**Box 2: Metacognition in Large Language Models (LLMs)**

Building on earlier precedents [91-94], interest in AI metacognition has exploded alongside LLMs [95].

Most work on LLM metacognition uses prompting. In chain-of-thought prompting, the model produces intermediate reasoning steps which are added to the model's context window to inform subsequent processing [96]. This is metacognitive in that the model is asked to interpret its reasoning processes and use that interpretation to improve its reasoning. Extensions to chain-of-thought prompting require more complex metacognitive control, such as chains that backtrack or branch ("tree of thought" [97] and "meta chain-of-thought" [98]), break problems into sequences of easier to harder subproblems ("least-to-most prompting" [99] and "plan-and-solve prompting" [100]), or consider multiple possible reasoning chains ("self-consistency" [101]). Other prompting techniques more explicitly incorporate wisdom-related metacognition. For example, metacognitive prompting [102] asks LLMs to consider several metacognitive queries at the same time as a primary query. Similarly, prompting a model to consider "could you be wrong?" leads it to consider data in its training set highlighting potential errors and biases [103].

Other approaches alter the module architecture by introducing additional modules or agents. The most common is a module for evaluating or monitoring proposed outputs. For example, EXAR uses a "meta-validator" module that assesses the outputs of a model and uses those assessments to fine-tune the model [104]. MIRROR separates a "Thinker" that creates an inner monologue and a "Talker" that uses the inner monologue as context for communicating to the user [105]. Another common addition is an explicit module or reasoning step for evaluating model capabilities, learning progress (e.g., in MAGELLAN [106]), or probability of task success (e.g., in MUSE [107]).

Still other techniques include mapping and evaluating the intermediate steps in chain-of-thought reasoning in a latent space [108], monitoring hidden layers for anomalies [109], comparing neural activation patterns in novel cases to baselines with known metacognition [110], treating metacognitive tasks such as self-awareness and self-evaluation as Bayesian learning problems [111], and incorporating "fast" and "slow" reasoning modes for routine versus complex problems [112].

Models with enhanced metacognition have been applied to a growing array of problems—few-shot classification [113], external tool selection [110], improving the use of external knowledge sources [114], inferring beliefs and desires [115], accurately classifying linguistic nuance such as sarcasm [116], and navigating simulated scenarios such as a zombie apocalypse [117]. Yet, these techniques remain imperfect and not all are wisdom-related (Figure 2).

## What are the potential benefits of wise AI?

We suggest that deficits in AI wisdom—and perspectival metacognition in particular—underlie failures in robustness, explainability, cooperation, and safety (Figure 2).

### *Robustness*

Given the range of intractable environments in which intelligent systems must operate, three failures of robustness are common: A system can be *unreliable*—given similar inputs, a system can produce wildly different outputs. This could be caused by applying different strategies each time, or applying a strategy that produces inconsistent results. A system can be *biased*—the output is systematically wrong or non-representative in a predictable direction. Or a system can be *inflexible*—novel inputs lead to lower-quality outputs.

Human wisdom combines object-level and metacognitive strategies to adapt robustly across environments. Object-level strategies like heuristics can be beneficial because they sometimes outperform analytic optimization by avoiding data overfitting [19], especially in novel, out-of-distribution contexts (but see [38]). These strategies are supported by wise metacognition, which helps reasoners to learn new information from other perspectives and discern its relevance, to balance the competing urges to simplify and optimize, and to avoid catastrophic error by checking the plausibility of a strategy's output.

For similar reasons, wise AI would be more robust in all three senses. It would be more reliable: Its monitoring processes would evaluate whether it is sensible to use different strategies in comparable situations and reject excessively inconsistent strategies. It would also be less biased: Since biased outputs usually result from biased inputs, a wise AI would reflect on its training data or models of the world, identifying sample deficiencies in its training data (perhaps requesting additional data), and understanding the causal process by which biases resulted (correcting for that bias). Finally, wise AI would be more flexible: It would moderate its confidence in novel situations, and would reduce, manage, and navigate uncertainty.

### *Explainability*

Opaque AI can produce puzzling outputs, difficult-to-diagnose errors, and barriers to collaboration [1]. Although cognitive scientists disagree about the extent of introspective access in humans [39], all theories agree that metacognition is necessary for justifying decisions to ourselves and others. Thus, wise AI would likely be more explainable.

One possibility is that, in humans, consciously accessible metacognitive strategies guide behavior. When we report our thought processes, we are reporting *observations*. For instance, the decision to moderate confidence in a prediction could be caused by a conscious recognition of ignorance, which can then be reported. The explainability problem is then reduced to selecting *which* of those observations to report, that is, which are the most relevant causal antecedents of the output.

On the opposite extreme, the mind may be “flat” [40]—it does not contain hidden depths of reasons that can be uncovered through introspection. When we report our thought processes, we report *inferences* (“stories”), not observations. The reasoner observes the outputs of her strategies and reasons backwards to their possible causes [41]. These inferences may often be incorrect [42], yet they are often useful justifications that, when expressed, constrain future thought and behavior. Since metacognition itself is not observable but only inferable, explainable AI would need to generate a useful narrative to make sense of its own actions—itself a metacognitive process.

Recent work suggests that even using techniques such as chain-of-thought [43] or metacognitive prompting [44] (Box 2), models confabulate insight rather than genuinely introspect, generating explanations not “faithful” to their underlying reasoning. Under the classical view, we would hope that techniques for improving the introspective accuracy of metacognition would yield more faithful explanations; if the mental flatness view is correct, all we can hope for is more useful post hoc reconstructions.

### *Cooperation*

AIs increasingly behave within larger networks, requiring both AI–AI cooperation (e.g., autonomous vehicles negotiating traffic) and AI–human cooperation (e.g., surgical robots), and influencing human–human cooperation (e.g., social media content curation). Cooperative AI [2,45] examines how AI can benefit all parties involved by navigating barriers to understanding, communication, and **commitment**. Wise object-level and metacognitive strategies are critical to how humans solve these problems, suggesting the same may be true for AI.

Cooperation requires understanding the social dynamics of the situation, including the likely actions taken by others. Since those actions depend on the beliefs and goals of agents, social understanding requires theory-of-mind [46], including the tacit ability to form joint plans to coordinate behavior [47]. In humans, this is accomplished through object-level strategies such as first-person simulation (putting oneself in the other’s shoes) [48] and third-person, theory-based reasoning (e.g., assuming that the agent is rational [49]).

Cooperation depends equally on communication—selecting and sending information to potential partners. Incoming information must be filtered to act on what is useful and ignore what is misleading or irrelevant [50]. Even young children develop object-level strategies for evaluating sources—tracking cues such as accurate past testimony and conflicts of interest [51]—and more sophisticated reasoners can check whether the reasoning itself is valid [52]. Such "epistemic vigilance" mechanisms make credible communication among humans possible: Without a means of assessing a communication, the risk of exploitation would undermine trust.

Cooperation can unravel when long-term incentives diverge, so humans have evolved ways to make credible commitments. Third-party social judgments—introducing potential punishment and reputational risk—impose external costs on defection [53], while emotions like shame and guilt impose internal costs [54]. Humans sharing a cultural and psychological context can assume these costs as common ground, promoting credible commitment.

Wise metacognition is required to effectively manage these object-level mechanisms [55-56]—resolving conflicts among strategies (e.g., when accuracy cues diverge), assessing their appropriateness (e.g., whether one can evaluate a chain of argumentation), and seeking appropriate inputs (e.g., knowing the capabilities of the other counterparty). This last point is particularly important for cooperative AI, which could overestimate the abilities of humans or lack common ground such as a shared emotional system.

***Safety***

Concerns about AI safety span the prosaic to the cataclysmic [3,57]. For now, the main safety risks are simply that systems that we come to rely on fail us—a shoddy surgical robot, incompetent tax advice, or biased parole algorithm. Machine metacognition can help to avoid such failures [58]. AIs with well-calibrated confidence can target the most likely risks; appropriate self-models would help AIs to anticipate failures; and continual monitoring of its performance would facilitate recognition of high-risk moments.

Some worry, however, that in the future, superintelligent machines will pose an existential risk to humanity if their goals are not 'aligned' with ours [59]. This concern arises from two observations: (i) Predefined goals are likely to be mis-specified or become obsolete, and (ii) a powerful AI could be difficult to curtail if it aggressively pursued the wrong goals. Bostrom [59] illustrates both points in his parable of the paperclip-maximizing AI that converts the Earth into paperclips and kills all humans in its way.

The goal of **AI alignment** [3] is to prevent such mismatches between the goals of an AI and its users—an exceedingly difficult task due to the many assumptions that are unspoken and potentially unshared. Wisdom is crucial to navigating such problems—first, because goal-specification is a prototypical example of an intractable problem for which we deploy wisdom; and second, because humans rely on 'common sense' wisdom to fill in such unspoken assumptions and make tacit agreements [60].

Indeed, we suspect that engineering wise social interaction—in addition to or perhaps instead of alignment—may be necessary to achieve alignment's goals. Alignment faces not only technical problems, but conceptual ones. *Who* should we align AI to? People differ in their goals (e.g., believing GenAI should solely aim to provide accurate information versus avoiding the reinforcement of harmful stereotypes) and values (e.g., cross-cultural and religious differences in maximizing happiness vs. liberty) [61]. Should we increase the average human well-being, its sum, or care for the whole biosphere? And why assume that today's values are the right ones, given profound shifts even over recent history [62]? Aligning AI to current values would risk reifying those values as "the right" values, stalling future social progress.

A two-pronged, wisdom-oriented approach may be more promising.

First, AIs must themselves implement wise reasoning—aligning them to the right object-level and metacognitive strategies rather than to the "right" values. For example, one object-level strategy may be a bias toward inaction (not executing an action if it risks harm according to one of several possibly conflicting human norms), which in turn requires metacognitive regulation (learning what those conflicting perspectives are and avoiding overconfidence).

Second, we must consider how AIs fit into a broader institutional ecosystem. Institutions like governments and markets address the 'alignment' problem that we humans have—ideally channeling our discrepant interests and values into socially productive directions. It is useful to think of AI not merely as an external tool influencing society but as a new type of agent within society, embedded in pairwise interactions and, increasingly, our broader institutions. If channeled effectively through institutions, metacognitively wise AI can enhance social evolution rather than undermine it. Both human and artificial agents in society should continue to allow our values to evolve toward a shared reflective equilibrium [63]—bringing situation-specific judgments and general moral principles into alignment with one another through iterative adjustments.

## How might we build wise AI?

Before considering how we can build wise AI, first consider how nature has built *us* to be wise. We suspect that metacognitive abilities are primarily evolutionary adaptations built into the architecture of the human brain, being fundamental across any context, whereas object-level strategies are primarily acquired through experience, including socialization and didactic learning, due to their great situational variability. While acknowledging the role of development, culture, and self-reflection for metacognition [64-65] and biological evolution for object-level strategies [66], we take "object-level = development" and "metacognition = evolution" as a starting point. If so, this suggests that implementing object-level and metacognitive wisdom may require different strategies.

In humans, object-level strategies like heuristics are typically acquired through trial-and-error and social learning. Since wise heuristics are often domain-specific, exhaustively specifying these rules is likely doomed for the same reasons that rule-based expert systems in AI failed. Instead, allowing AI systems to learn from experience [67] and from others [68] may be more promising.

The analogy to the human case suggests, however, that experience alone is unlikely to suffice for training metacognition. One problem is that in typical AI training, a loss function is minimized, which is defined over the model's outputs rather than its reasoning. Although this may indirectly select for sound decision-making strategies, the poor explainability of many state-of-the-art models makes it difficult to determine what those strategies are; an output may please a human judge for the wrong reasons. Such a system would often emulate the decisions of a wise human, but would not itself be metacognitively wise.

How might one get around this problem? Optimistically, standard LLM training techniques could be modified. For example, a two-step training process could be implemented in which a model is first trained for wise strategy selection directly (e.g., correctly identifying when to be intellectually humble) and then training them to use those strategies correctly (e.g., carrying out intellectual humble behavior). Alternatively, one could present models with benchmark cases, request them to produce both their metacognitive strategy and their output, and then reward only the correct combination of strategy and output [69]. In either case, models could be trained against what a wise human would do or against the acceptability of its explanations for its choices.

Perhaps, however, no amount of training will get current models to human-level metacognition, just as no amount of language exposure will get a squirrel to talk. On this view, the "innate" architecture of current models is not up to the task. LLMs work by generating the next token (i.e., word or word part) based on the input in its **context window**. At first, this input comprises the user's prompt; after the model is run to generate the first token in its response, this token is added to the context window, and the model is re-run to generate the second response token, and so on. This process does not involve feedback from later layers to earlier ones and it is backward-looking—it predicts one word ahead based on its input and output-so-far, rather than explicitly planning ahead. This process can yield surprisingly intelligent outputs—and even some degree of planning (e.g., rhyming in a poem [70])—given enough parameters and data. Yet, given their lack of explicit planning, perhaps it is unsurprising that LLMs struggle with metacognition, which requires reflecting on one's thoughts and devising strategies to regulate them. Changes to model prompting and architecture may be required, not just changes to training. Box 2 describes some ongoing efforts in this spirit, while Table 1 lists some more speculative ideas.

**Table 1: Engineering Wiser AI via Metacognition**

| Conceptual idea | Possible implementations |
|---|---|
| **1. Explicit metacognitive checkpoints and error detection loops**<br><br>Integrate explicit reflective checkpoints into AI decision-making processes, forcing the AI to periodically evaluate coherence, reliability, and confidence in its reasoning. Implement continuous error detection loops where an AI system revises internal strategies upon encountering prediction failures or contradictions. | Introduce specific computational modules at defined decision points (e.g., transformer layers in LLMs) that assess output uncertainty (entropy, calibration error) and coherence metrics (consistency with past outputs).<br><br>Implement error detection using confidence thresholds learned from validation data. For instance, pause execution to reassess decisions whenever model confidence falls below calibrated uncertainty thresholds, forcing conditional re-generation or seeking external verification. |
| **2. Epistemic source tagging and reliability updating**<br><br>Implement structured metadata that explicitly encodes epistemic reliability for training data sources. Allow systems to dynamically update their trust in data sources (provenance and lineage) based on consistency of predictions and feedback, akin to human epistemic vigilance mechanisms. | Precompute and embed metadata vectors capturing reliability indicators (e.g., historical accuracy, domain expertise scores, publication credibility metrics) alongside raw tokens or data points.<br><br>Train AI systems to dynamically adjust reliability scores using a simple online Bayesian updating mechanism: sources whose information frequently results in erroneous outputs or internal contradictions receive lowered reliability scores, reducing their influence during inference. |
| **3. Hierarchical and reflective reasoning architectures**<br><br>Employ hierarchical architectures inspired by cognitive models (e.g., ACT-R [118], SOAR [119]), where a metacognitive layer explicitly monitors and selects object-level strategies. Develop explicit reflective subsystems designed to audit internal consistency and logical coherence of reasoning outputs, promoting effective “sanity checking.” | Implement cognitive-architecture-inspired hierarchical models, using explicit controller modules (meta-policy networks) to govern lower-level task-specific modules: a) Hybrid symbolic/sub-symbolic approaches (e.g., OpenCog Hyperon [120], ACT-R style modules); b) Reinforcement learning hierarchical controllers (e.g., FeUdal networks [121])<br><br>Introduce standalone “auditor” modules trained explicitly to critique primary outputs for internal consistency, logical coherence, or sensitivity to constraints. For instance, chain-of-thought prompting [96] or future advanced reasoning modules explicitly trained as reasoning auditors. |
| **4. Transparency via metacognitive narration**<br><br>Design systems capable of transparently narrating their internal metacognitive reasoning (“thinking aloud” protocols) to users, aiding explainability and making reasoning easier to audit and debug. | *“Thinking Aloud” protocols:* Implement explicit model training on explanatory datasets or devise new chain-of-thought approaches, which generate explicit narration of metacognitive reasoning steps in understandable language.<br><br>*Interactive debugging & auditing interfaces:* Build interactive visualization tools displaying model uncertainty, reasoning trails, or decision checkpoints to users or system auditors. |
| **5. Distributed and social metacognition** | *Multi-agent epistemic vigilance:* Multiple independent AI agents work collaboratively, requiring agreement or consensus for outputs on |
| Leverage multi-agent reasoning and collective decision-making, analogous to human reliance on socially distributed cognition. Implement epistemic cross-checking and adversarial debate between multiple AI systems to mitigate individual AI overconfidence and misinformation propagation. | critical tasks. *Concrete architectures:* Multi-agent RL (MARL) [122], decentralized autonomous organizations (DAO)-inspired decision-making [123].<br><br>*Debate-based metacognitive cross-checking:* AI reasoning outputs must pass adversarial debates or cross-examinations from independently trained AI debaters before being finalized. *Example frameworks*: OpenAI's debate-style AI safety approach [124], Anthropic's Constitutional AI approach [125]. |
| **6. Scheduled off-line replay & consolidation**<br><br>Use off-line periods for AI systems to consolidate and "reflect" on prior model runs, akin to one possible function of the human default mode network [65,126] | Allocate compute to periods during which outward action pauses while the model regenerates latent trajectories, pits alternative chains of thought against each other (self-consistency / debate), and refreshes its calibration curves before the next on-line cycle. |

### *Evaluating Machine Wisdom*

Once we build a wise machine, how will we know it? Wisdom is context-sensitive, so a **benchmark** input must contain sufficient detail to match the rich context of a real-world situation. Moreover, since wisdom is about the reasoning underlying strategy selection, any evaluation procedure must judge not only the outcome but the precipitating process.

Existing benchmarking work in metacognition has focused on the calibration of confidence judgments [29,71]. An advantage of this narrower domain is that it is much more tractable than the perspectival metacognitive strategies we have discussed here, with well-developed methods that even work in non-human animals [72], lend themselves to computational modeling [73], and are able to separate performance on the cognitive versus metacognitive component of a task [71,74]. Nonetheless, these tasks are domain-specific, often constrained to well-defined laboratory environments, and do not yet capture the richness of everyday intractable problems that wise judgment handles.

To make progress, let's consider how other rich, complex constructs have been benchmarked. One approach is to collect tasks from psychology experiments, akin to benchmarking theory-of-mind or analogical reasoning [75-76]. Since these tasks are discussed in the literature (and appear in training data), the content must be replaced with structurally similar but superficially different problems [77-78]. However, since these tasks usually measure outcomes only and provide little context, this approach cannot be adopted wholesale for wisdom. An alternative approach—used to benchmark explanatory abilities [79]—is for domain experts to subjectively evaluate the quality of the model's outputs. This approach is well-suited for evaluating reasoning (rather than outcomes), but requires some form of quantification to compare models.

One way to evaluate AI wisdom would start with tasks that measure wise reasoning in humans [80]. These tasks present participants with a social dilemma or a choice between seemingly incommensurable options, asks them to reflect on the next steps, with reflections scored on prespecified criteria by human raters, such as experts. Novel and detailed variants of such scenarios could be presented to AIs, with their performance scored by either human raters or by other models (if their scores converge) [81]. It would be important to include problems that agentic AIs might confront in the future (e.g., whether to execute a debatably ethical request), to ensure they can reason wisely not only about humans but about themselves.

Ultimately, the wisdom of increasingly autonomous AIs will be judged by human users and stakeholders. Prior benchmarking is a crucial start, but there is no substitute for interacting with the real world. Given this intrinsic limit on our ability to evaluate wisdom *ex ante*, this integration with the world must proceed slowly to minimize risks.

## Concluding Remarks

Building smarter machines comes with risks: AI with advanced capabilities might pursue undesirable goals. Is there a parallel concern about the unintended consequences of building wiser machines?

Perhaps not. Empirically, humans with wise metacognition show greater orientation toward the common good, including cooperation and responsiveness to others [55] (Box 1). Perhaps wise AI would have these qualities too.

Yet, an important ambiguity arises here about which we can only speculate: Although current AI may not be wise, what shape would a future AI's wisdom take?

One possibility is that AI and human wisdom might sharply diverge. Human metacognition serves largely to economize scarce cognitive resources [82-83], and many biases may be side-effects of solving this constrained optimization problem [84-85]. Given the more abundant computational resources of wise AI, this optimization problem may look very different from humans'—AIs might rationally invest far more effort. Conversely, humans outsource much of our cognition to the social environment (as in the division of physical or cognitive labor [86-87]), including knowledge-generating institutions that are ever-evolving. Distributed cognition of this sort is not yet a dominant paradigm in AI and it is unclear what its (dis)advantages are compared to an extensive, integrated knowledge base.

Conversely, perhaps AI wisdom would converge considerably with human wisdom. AI wisdom also faces computational constraints, since compute can be costly. Moreover, heuristics work for AI for the same reasons as they work for humans: When we lack complete information, heuristics can perform well by implementing sensible, robust defaults. Finally, AIs may come to join our social environment—and perhaps reap some of the same social cognitive advantages as humans—as AI is increasingly integrated into human institutions [88].

Given these considerations, uncertainty remains (see Outstanding Questions). What if we tried and failed to build wise AI? What if the characteristics of wise AI differ from those of a wise human, to the detriment of humans? To these concerns we have three responses.

First, if the alternative were halting all AI progress, building wise AI would introduce added risks. But compared to the status quo—advancing capabilities at a breakneck pace without wise metacognition—the attempt to make machines intellectually humble, context-adaptable, and adept at balancing viewpoints seems clearly preferable.

Second, at least in the medium term, AI will not act autonomously but will remain a collaborative tool to be used by and for humans, supporting rather than replacing human wisdom. In this sense, understanding how humans and AIs might work together to produce wise or foolish decisions becomes a crucial research agenda.

Finally, the qualities of robust, explainable, cooperative, and safe AI will amplify one another. Robustness facilitates cooperation (improving confidence from counterparties) and safety (avoiding failures in novel environments). Explainability facilitates robustness (aiding human intervention through transparency) and cooperation (more effective communication). Cooperation facilitates explainability (accurate theory-of-mind about users) and safety (implementing shared values). Wise metacognition can lead to a virtuous cycle in AI, just as it does in humans. We may not know precisely what form wise AI will take—but it must surely be preferable to folly.

## Glossary

- **AI alignment:** Ensuring that AIs pursue the goals intended by (“aligned with”) their human users.
- **benchmark:** A set of standard tasks on which AIs can be compared to one another and to humans for a given capacity.
- **commitment:** The ability to make a credible promise that will be kept at a later time, particularly as a means of incentivizing a mutually beneficial cooperative agreement.
- **context window:** The sliding window of text that a GenAI model has access to (can “remember”) when formulating its output.
- **conflict resolution process:** A type of metacognitive process that selects the best strategy when object-level strategies conflict.
- **cooperative AI:** AI that is able to pursue shared goals—with other AIs or with human users—through abilities including social understanding, communication, and credible commitment.
- **decision technologies:** Organized procedures for making decisions, such as formal calculation.
- **explainable AI:** AI that can be effectively understood by users, for instance because the AI can effectively communicate its decisions and reasoning to users.
- **heuristic:** An object-level strategy that produces a solution to a problem without conducting a full analysis, typically by using a subset of the available information.
- **input-seeking process**: A type of metacognitive process that seeks the inputs required for object-level strategies to work.
- **intractable problem:** A problem that does not lend itself to analytic techniques such as optimization.
- **metacognitive strategy:** A strategy that is used to manage other (especially object-level) strategies, including by seeking the required inputs, resolving conflicts among strategies, and monitoring the plausibility of outcomes.
- **narrative thinking:** An object-level strategy in which an individual constructs a causal and analogical model of a situation in order to understand a situation, predict how it will unfold, and evaluate potential choices.
- **object-level strategy:** A strategy that is used to produce a potential solution to a specific problem or task, such as a heuristic, narrative, or analytic procedure.
- **outcome-monitoring process:** A type of metacognitive process that checks whether outcomes of the selected object-level strategy are plausible (also called “sanity checking”).
- **perspectival metacognition:** A subset of metacognitive skills for managing and integrating perspectives on a situation.
- **robust AI:** AI that works effectively in novel environments because it is reliable (similar inputs yield similar outputs), unbiased (not systematically mistaken), and flexible (able to generalize to novel inputs).
- **safe AI**: AI that avoids risks associated with harmful failures, which can include both incompetence (e.g., errors because the AI is not robust) or malevolence (e.g., malfeasance because the AI is not aligned).

- **wisdom:** A suite of abilities used to solve intractable problems, comprising both metacognitive strategies (e.g., intellectual humility) and object-level strategies (e.g., culturally transmitted heuristics).

## Outstanding Questions

- How might wise AI inform—and be informed by—the cognitive science of human wisdom? For instance, how can computational modeling of human wisdom (including object-level and metacognitive strategies) and efforts to engineer machine wisdom be mutually enlightening?
- What is the best approach to formalizing wise reasoning in mathematical approaches to AI robustness, explainability, cooperation, and safety?
- How might the give-and-take of conversation between humans and AI lead to a form of shared wisdom? How should this potential for collaborative metacognition inform the design of AI systems?
- Might AI wisdom exceed human wisdom? If so, how would we humans know?
- How would the mass adoption of wise AI impact society? For example, could this lead to offloading of metacognitive labor, leading to a decline in human wisdom? Or could wise AI act as a cognitive prosthetic to enhance human wisdom in practice?
- Could wise AI be subverted to malicious ends? Might wiser AI counter this problem, or exacerbate it?
- What can we learn about existential AI risks by studying wise and unwise human decision-making and institutional design around other existential risks such as nuclear weapons?
- Where would AI not benefit from wise metacognition—for instance, because the benefits are marginal relative to economic, environmental, or computational costs?
- How would metacognitive AI systems scale up? How would the further integration of wise AI into human institutions impact the functioning of those institutions and of AI itself?
- What further considerations would be required to embody metacognition in robots?